%% file: main.tex
\documentclass[a4paper, 10pt, conference]{ieeeconf}      
                                          
\usepackage{tikz}                
\usepackage{FG2026}
\usepackage{graphicx}
\usepackage{subcaption}
\usepackage{booktabs}
\usepackage{framed}
\usepackage{amsfonts}
\usepackage[table]{xcolor} 
\FGfinalcopy 

\DeclareFontFamily{U}{FdSymbolA}{}
\DeclareFontShape{U}{FdSymbolA}{m}{n}{
    <-> s * [1] FdSymbolA-Book
}{}
\DeclareFontShape{U}{FdSymbolA}{m}{b}{
    <-> s * [1] FdSymbolA-Medium
}{}
\DeclareSymbolFont{fdsymbols}{U}{FdSymbolA}{m}{n}
\SetSymbolFont{fdsymbols}{bold}{U}{FdSymbolA}{m}{b}

\DeclareMathSymbol{\medtriangleup}{\mathbin}{fdsymbols}{83}
\DeclareMathSymbol{\medtriangledown}{\mathbin}{fdsymbols}{85}

\newcommand{\updegrade}{\textcolor{red}{\medtriangleup}}
\newcommand{\downimprove}{\textcolor{darkgreen}{\medtriangledown}}
\newcommand{\nochange}{\textcolor{gray}{\mathbf{\sim}}}

\IEEEoverridecommandlockouts                              
\usepackage{amsmath} 
\usepackage{multirow}
\usepackage{xcolor}
\def\FGPaperID{215} 
\usepackage{hyperref}

\title{\LARGE \bf
Investigating Bias and Fairness in Appearance-based Gaze Estimation
}

\author{\parbox{16cm}{\centering
    {\large Burak Akgül$^1$, Erol Şahin$^1$, and Sinan Kalkan$^1$}\\
    {\normalsize
    $^1$Dept. of Computer Eng. and ROMER Robotics Center,  Middle East Technical University, Ankara, Turkey}}
}

\newcommand\copyrighttext{%
  \footnotesize \textcopyright 2026 IEEE. Personal use of this material is permitted.  Permission from IEEE must be obtained for all other uses, in any current or future media, including reprinting/republishing this material for advertising or promotional purposes, creating new collective works, for resale or redistribution to servers or lists, or reuse of any copyrighted component of this work in other works.}
\newcommand\copyrightnotice{%
\begin{tikzpicture}[remember picture,overlay]
\node[anchor=south,yshift=10pt] at (current page.south) {\fbox{\parbox{\dimexpr\textwidth-\fboxsep-\fboxrule\relax}{\copyrighttext}}};
\end{tikzpicture}%
}

\begin{document}

\definecolor{darkred}{RGB}{139,0,0}
\definecolor{darkgreen}{RGB}{0,100,0}

\ifFGfinal
\thispagestyle{empty}
\pagestyle{empty}
\else
\author{Anonymous FG2026 submission\\ Paper ID \FGPaperID \\}
\pagestyle{plain}
\fi
\maketitle
\copyrightnotice

\begin{abstract}

While appearance-based gaze estimation has achieved significant improvements in accuracy and domain adaptation, the fairness of these systems across different demographic groups remains largely unexplored. To date, there is no comprehensive benchmark quantifying algorithmic bias in gaze estimation. This paper presents the first extensive evaluation of fairness in appearance-based gaze estimation, focusing on ethnicity and gender attributes. We establish a fairness baseline by analyzing state-of-the-art models using standard fairness metrics, revealing significant performance disparities. Furthermore, we evaluate the effectiveness of existing bias mitigation strategies when applied to the gaze domain and show that their fairness contributions are limited. We summarize key insights and open issues. Overall, our work calls for research into developing robust, equitable gaze estimators. To support future research and reproducibility, we  publicly release our annotations, code, and trained models at: 
\url{github.com/akgulburak/gaze-estimation-fairness}
\end{abstract}


\section{INTRODUCTION}

Gaze provides a fundamental window into human cognitive status, mental health, and social dynamics \cite{gaze_estimation_survey}. Analyzing this vital signal effectively can bring in information about human intention \cite{gaze_importance_survey}, attention \cite{review_driver_gaze_behavior_understanding}, and communication patterns \cite{bayesian_gaze_estimation_hri_intro}, which can potentially enrich many applications with human subjects. 

Accurate estimation of gaze has attracted significant interest in the literature (see, e.g., \cite{appearance_based_gaze_survey, automatic_gaze_survey, eye_of_beholder_survey_gaze, gaze_estimation_survey, gaze_importance_survey} for reviews). 
Gaze can be estimated using two main approaches: Model-based or appearance-based methods \cite{gaze_estimation_survey}.
Model-based methods use the eyeball structure and the geometric model of the eye to estimate the gaze \cite{eye_of_beholder_survey_gaze}. On the other hand, appearance-based methods directly map images to human gaze and do not require the use of geometric features \cite{appearance_based_gaze_survey}. Compared to the model-based methods, appearance-based methods are shown to be more robust \cite{eye_of_beholder_survey_gaze}, and have low compute requirements. Consequently, appearance-based methods are more commonly used for gaze estimation \cite{appearance_based_gaze_survey}. 



\begin{figure}
    \centering
    \includegraphics[width=0.99\linewidth]{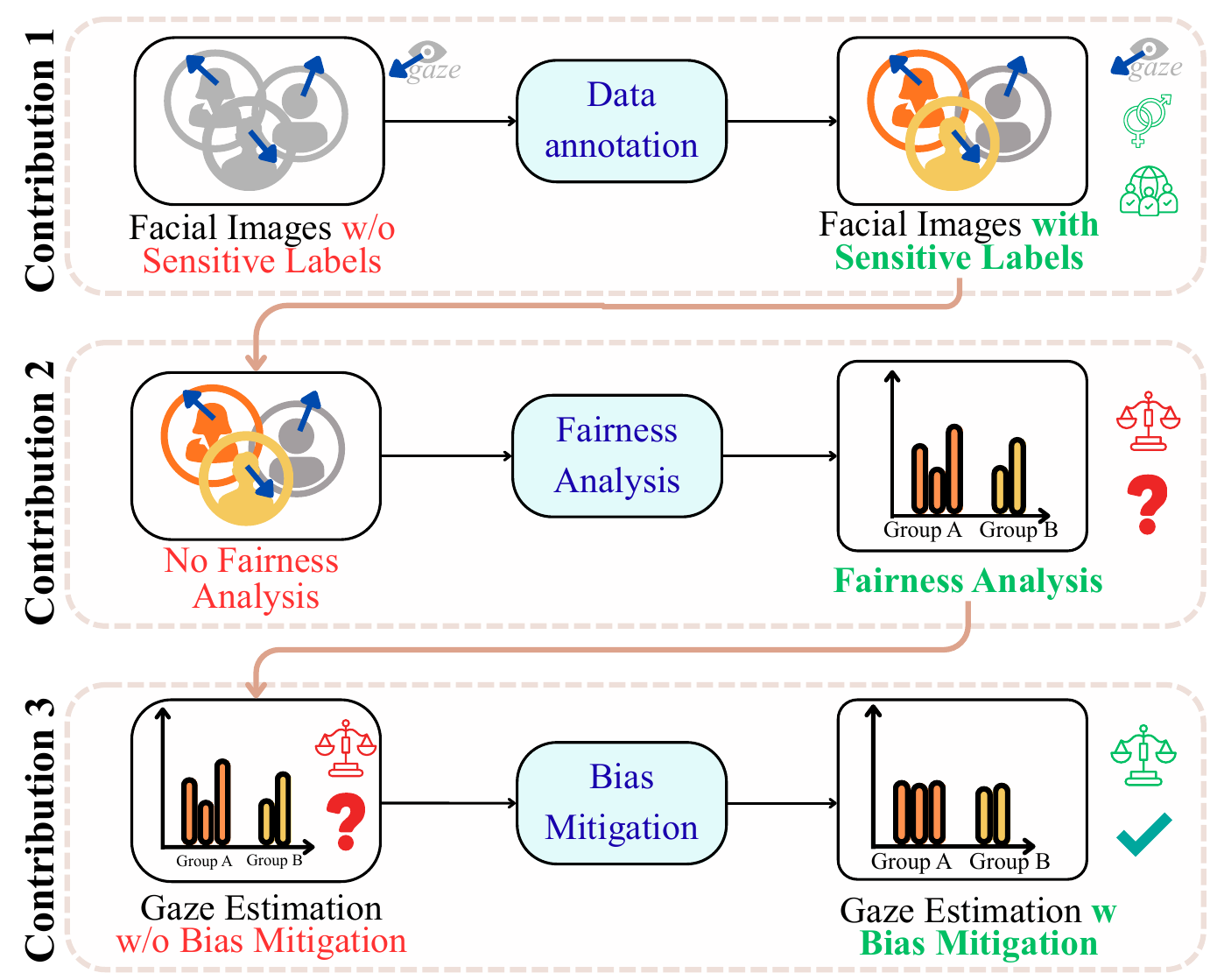}
    \caption{Contributions of the paper for fairness in gaze estimation. Contribution 1: Annotating the selected datasets with sensitive attributes. Contribution 2: Analysis of fairness between demographic groups, using the sensitive attribute annotations. Contribution 3: Applying bias mitigation methods to reduce bias and achieve fairness.}
    \label{fig:teaser}
\end{figure}

Machine learning models are known to suffer from biased predictions across demographic groups to which facial signal processing is no exception. Prior work extensively showed that many facial signal processing problems, e.g., face recognition \cite{yucer2024racial}, face detection \cite{menezes2021bias}, facial expression recognition \cite{hosseini2025faces}, facial affective computing \cite{cheong2021hitchhiker}, do include various forms of biases towards underrepresented groups.  

\textbf{Research Gaps}. Despite the versatility of gaze estimation across different domains and applications \cite{fair_sa_fairness_face_recognition, mixfairface_face_recognition, consistent_fairness_face_recognition,enhance_fairness_facial_signal_process,masked-face-recognition} and the fairness issues commonly reported in many facial signal processing applications \cite{cheong2021hitchhiker,hosseini2025faces,menezes2021bias,yucer2024racial}, bias and fairness in appearance-based gaze estimation methods have not received any attention in the literature (\textbf{Gap 1}). This is concerning because many high-stake applications, e.g., driver monitoring systems, can lead to life threatening outcomes owing to biases across demographic groups. Furthermore, there is no study focusing on mitigating the bias issues in gaze estimation (\textbf{Gap 2}). One fundamental reason for these two gaps is the lack of demographic labels for the currently published gaze estimation datasets (\textbf{Gap 3}), where only the gaze label is available.


\textbf{Contributions}. 
In this paper, we address the aforementioned research gaps ({Gap 1-3}) with the following contributions (see also Figure \ref{fig:teaser}):
\begin{itemize} 

    \item \textbf{Sensitive attribute annotation.} We annotate existing commonly-used gaze estimation datasets, namely Gaze360 and GazeCapture \cite{gaze360,gazecapture}, with the gender and ethnicity sensitive attributes. To foster research into the analysis and mitigation of any fairness issues in the literature, we will publicly share these annotations.

    \item \textbf{Bias analysis in gaze estimation.} We systematically analyze bias in gaze predictions of existing state-of-the-art gaze estimation models (namely, CrossGaze \cite{crossgaze}, MCGaze \cite{mcgaze}, L2CS-Net \cite{l2csnet}, PureGaze \cite{puregaze}, GazeTR \cite{gazetr}). Our analysis highlights severe fairness issues across ethnicity and gender.

    \item \textbf{Bias mitigation.} We form a baseline benchmark for fairness algorithms by evaluating several pre-processing and in-processing bias mitigation algorithms. We demonstrate that standard bias mitigation techniques provide limited effectiveness for gaze regression, highlighting the need for geometric or domain-specific solutions. 
    
    \item \textbf{Highlighting Open Issues.} Overall, our work highlights open issues and calls for research into developing more robust, equitable gaze estimators as well as designing more effective bias mitigation strategies.

\end{itemize}

\section{Literature Review}

\subsection{Gaze Estimation Methods}
Gaze estimation methods are commonly grouped under two categories: Model-based methods and appearance-based methods. Even though model-based methods achieve good accuracy they need personal calibration and require specialized hardware, such as depth sensors \cite{model_based_tof_camera}, infrared cameras \cite{model_based_rgb-ir_cameras}, and light sources \cite{model_based_gaze_tracking_single_light_source}. On the other hand, appearance-based methods have low device requirements, where a simple rgb camera can be used.

\noindent\textbf{Model-based Gaze Estimation.} 
Model-based gaze estimation methods are divided into two main categories, 2D eye feature regression methods \cite{learning_to_find_eye_region} and 3D eye model recovery methods \cite{eye_model_rgbd_camera}. 2D eye feature regression methods use features from the eye-region to find the gaze direction. Martinez et al. \cite{non_linear_regression_2d_eye_feature} infer the gaze with multilevel HOG features. Wood et al. \cite{eyetab_model_based} determine eye ROI and limbus outline to find the gaze direction without using the optical axis. On the other hand, 3D eye model recovery methods construct 3D structure and pose of the eye to calculate the gaze direction. Wang et al. \cite{model_based_3d_deformable_eye_face_model} uses kinect sensor and create 3D deformable eye-face model. Wood et al. \cite{3d_morphable_gaze_estimation} use multi-part eye region with 3D morphable model with analysis-by-synthesis.


\noindent\textbf{Appearance-based Gaze Estimation.} 
Appearance-based methods directly map eye appearance to human gaze. Existing methods can be grouped by analyzing their feature extraction method with respect to the input type, the deep network architecture that they utilize and the calibration step for personal variances or domain shifts \cite{appearance_based_gaze_survey}. 

As the first step, representative features are extracted from the input, nowadays using deep pretrained network. Gaze estimation inputs are categorized as face images, eye images, and videos. Some methods solely rely on the usage of eye regions for gaze estimation \cite{asymmetric_regression_eye_image, mobile_devices_eye_image, differential_approach_eye_image,deep_pictorial_eye_image}. 
In addition to using eye images, utilizing face images improves gaze accuracy \cite{l2csnet,efe}. 
And finally, some methods also utilize temporal information between video frames \cite{mcgaze,gaze360}. 

Gaze inputs are processed by variety of deep learning architectures in appearance-based gaze estimation. CNNs are the most widely used architecture in the literature for gaze estimation \cite{l2csnet,crossgaze,puregaze,canet}. Transformers, which capture global relationship better than CNNs, are also studied \cite{gazetr}. Recurrence-based architectures that utilize temporal information between frames are used with video inputs \cite{mcgaze,gaze360}.

In addition to using different architectures, personal calibration and domain adaptation methods also improve accuracy of the estimated gaze. Personal calibration improves subject-wise gaze accuracy \cite{offset_calibration,low_complex_calibration}, while domain adaptation methods increase transferability of methods between different domains \cite{puregaze, pnp-ga+, pnp-ga, crga}.

\subsection{Bias and Fairness in Machine Learning}

Measuring and mitigating bias to obtain fairer machine learning predictions has been widely studied in the machine learning literature. Fairness is generally measured by looking at statistical disparities among individuals \cite{two_ways_individual_fairness} or groups of individuals \cite{disparate-impact, demographic_parity, statistical_parity}.

To mitigate bias, the literature has proposed a plethora of approaches which can be broadly analyzed as pre-processing, in-processing, or post-processing methods (see, e.g., \cite{bias_mitigation_methods_survey} for reviews). Pre-processing techniques are applied to the training data to obtain a balanced distribution across demographic groups before using the data to train a model \cite{reweight, fairness-harm, mixup}. In-processing bias mitigation techniques modify the architecture, the objective or the training dynamics to prevent disparate processing of inputs belonging to different demographic groups \cite{fairness-adaptive-weights, fairness-aware-classifier, adversarial-debias}. Post-processing methods, on the other hand, are applied to the predictions of the model after its training is complete \cite{equal-opportunity, calibrated-equalized-odds}. The bias mitigation techniques are generally used for classification problems and there is little work that deals with bias in regression problems \cite{bagging_boosting_bias_regression}.

\subsection{Bias and Fairness in Facial Signal Processing}


Bias and fairness problem has been studied within the domain of face recognition, face verification, emotion/expression recognition, or face and gesture detection \cite{faces_of_fairness,fair-sa,bias_analysis,analysis_demographic_biases,investigating_bias_fairness}. These studies have shown that deep networks are very sensitive to changes across color, illumination, gender-related and age-related attributes in facial images, which manifest as significant disparities among demographic groups across different applications using facial images. 
Following the machine learning fairness literature, several pre-processing, in-processing and post-processing bias mitigation methods have been explored for facial bias \cite{fairness_calibration,mitigation_skewness_aware,investigating_bias_fairness}. 




\subsection{Comparative Summary}

As reviewed above, despite their severe implications, bias and fairness issues in appearance-based gaze estimation methods have not been addressed in the literature. In our work, we fill this gap by analyzing bias across several datasets and mitigating any bias issues using pre-processing and in-processing approaches.

\section{Method}

\begin{figure}[h!t]
    \centering
    \includegraphics[width=1.0\linewidth]{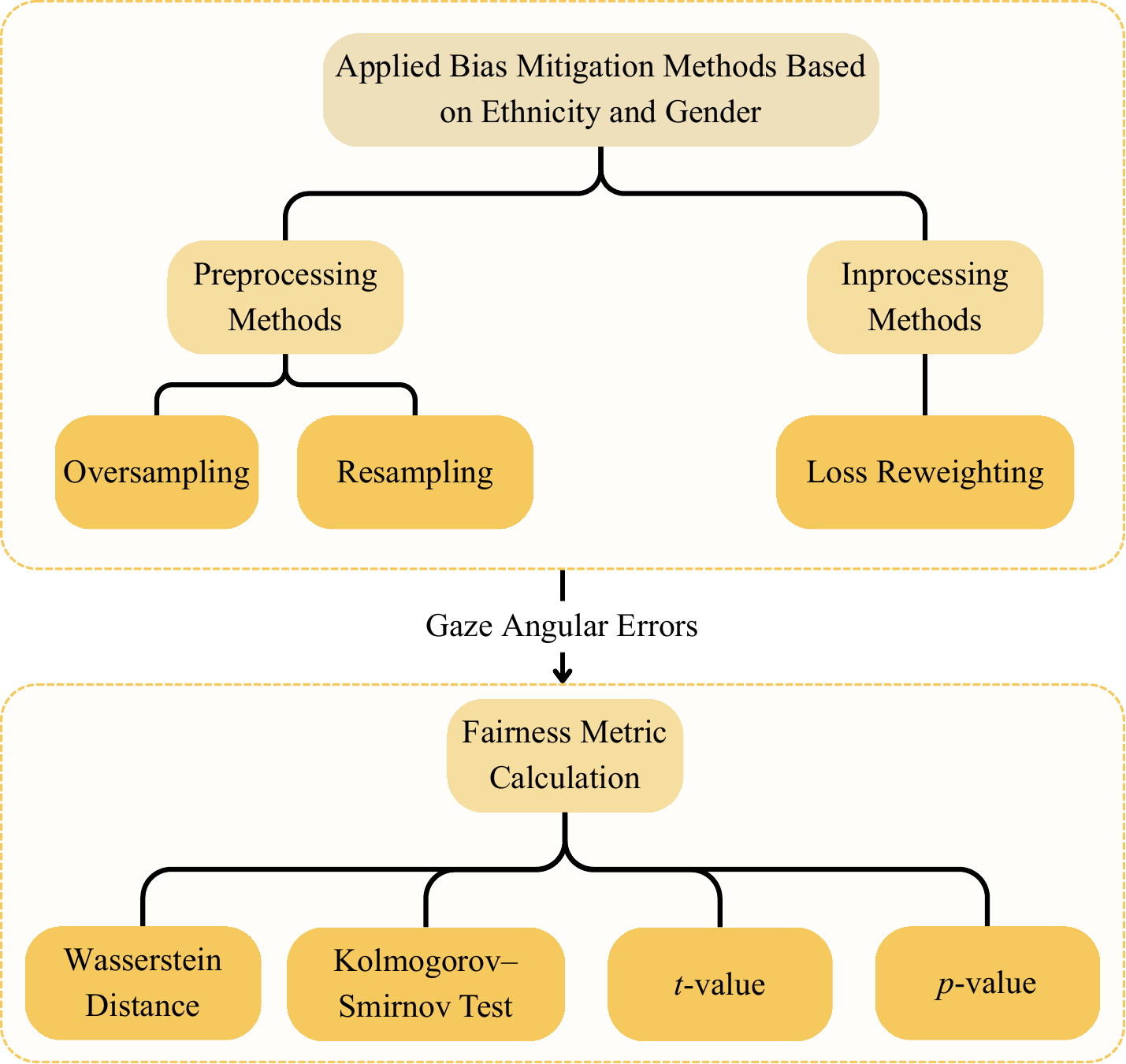}
    \caption{Overall pipeline for bias mitigation methods and fairness calculations.}
    \label{fig:labelingPipeline}
\end{figure}

Gaze estimation datasets are not provided with ethnicity or gender labels. These protected attributes are needed for analyzing fairness and applying bias mitigation methods. The labeling procedure is explained in the data annotation section.

Bias mitigation methods are applied to the ethnicity and gender labeled dataset Gaze360. Details of the bias mitigation methods are explained further in the bias mitigation methods section.

Analysis and evaluation of the selected models are calculated for the Gaze360 \cite{gaze360} and the GazeCapture \cite{gazecapture} datasets. Since the GazeCapture dataset is not provided with the 3D gaze annotations, the method described in \cite{l2csnet} is used to convert the labels.

\subsection{Problem Definition and Notation}

Gaze estimation is commonly formulated as a regression problem. Let $\mathcal{D} = \{(\mathbf{x}_i, \mathbf{y}_i, s_i)\}_{i=1}^N$ denote the dataset, where $\mathbf{x}_i \in \mathbb{R}^{H \times W \times C}$ is the input image, $\mathbf{y}_i \in \mathbb{R}^2$ represents the ground truth gaze angles $(\theta_{yaw}, \phi_{pitch})$, and $s_i \in \mathcal{S}$ is the sensitive attribute (e.g., gender or ethnicity). We define a function $F(\mathbf{y})$ that converts gaze angles to a unit 3D direction vector $\mathbf{v} \in \mathbb{R}^3$:
\begin{equation}
    \mathbf{v}_i = F(\mathbf{y}_i) = 
    \begin{bmatrix} -\cos(\theta)\sin(\phi) \\ -\sin(\theta) \\ -\cos(\theta)\cos(\phi) \end{bmatrix},
\end{equation}

We use a deep network $f(x_i; \Theta)$ with parameters $\Theta$ to predict the gaze angles $\hat{\mathbf{y}}_i$. We calculate the angle between the predicted gaze vector $\hat{\mathbf{v}}_i = \hat{\mathbf{y}}_i$ and ground truth gaze ${\mathbf{v}}_i = F(\mathbf{y}_i)$ as the error $\ell_i$ the model makes: 
\begin{equation}\label{eq:error}
    \ell_i = \arccos\left( \frac{\mathbf{v}_i \cdot \hat{\mathbf{v}}_i}{\| \mathbf{v}_i \| \| \hat{\mathbf{v}}_i \|} \right).
\end{equation}



\subsection{Sensitive Attribute Annotation}
\label{sect:data_annotation}


\begin{figure}[h!t]
    \centering
    \includegraphics[width=1.0\linewidth]{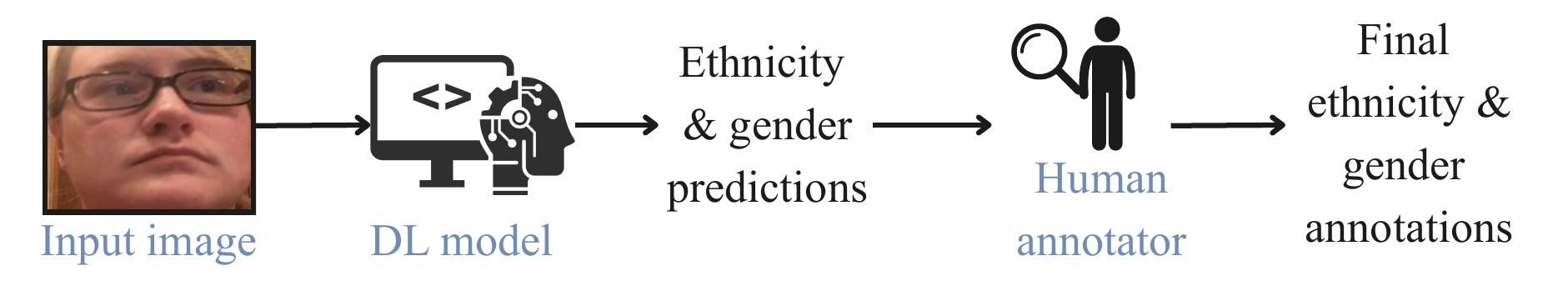}
    \caption{The ethnicity and gender annotation pipeline. A deep learning model (Fairface \cite{fairface}) is used to annotate the subjects. Conflicting annotations are validated by the human annotator.}
    \label{fig:labelingPipeline}
\end{figure}


The commonly-used gaze estimation datasets (Gaze360 \cite{gaze360} and GazeCapture \cite{gazecapture}) do not include demographic group labels. Therefore, we annotate ethnicity and gender labels for those datasets. We labeled the datasets with three ethnicity labels, Caucasian, Asian, and Afro-American, and two gender labels, male and female. 

We used the following procedure for annotating the sensitive labels for both Gaze360 and GazeCapture: 
We first used the Fairface \cite{fairface} model, which estimated ethnicity and gender labels from face crops of each subject. Then we validated the labels with a human annotator. If the model makes an apparent mistake, the human annotator changes the subject label. If the human annotator is not confident about whether the model output is correct or not, the subject's ethnicity is labeled as Unsure. 

The GazeCapture dataset contains 1474 recording sessions.
The Gaze360 dataset has 715 (recording session, person identity) pairs, where each pair corresponds to a subject. The person identity is not unique, and different person identities can correspond to the same subject. Thus, the labeling procedure is conducted on the recording session - person identity pairs. 


\subsection{Bias Mitigation Methods}
\label{sect:bias_mitigation}

To address bias in gaze estimation, we consider pre-processing \cite{maximum-entropy, reweight, fairadapt_pre_process_regression, oversample_fairness_visual_recognition} and in-processing \cite{adversarial-debias, model_correction_in_process_regression, fairness_class_weight} methods that have been proposed in the literature for addressing bias in regression problems. Pre-processing bias mitigation methods are applied to data before training, where we select oversampling and resampling. In-processing methods are applied during training, where we use loss reweighting. 

Post-processing bias mitigation methods are applied to calibrate model outputs without changing the model architecture or training the model again \cite{model_agnostic_post_process_regression,individual_fairness_post_process_regression}. However, gaze estimation cannot be calibrated by using sensitive attributes only -- the image input must also be used for calibration. To the best of our knowledge, such a method has not been proposed in the literature. Since proposing a new postprocessing method to deal with this issue is beyond the scope of this paper, post-processing methods are not analyzed.


\textbf{Pre-processing: Oversampling}. Oversampling methods modify the sampling probabilities for different demographic groups to mitigate bias \cite{oversample_fairness_visual_recognition}. More formally, the probability to select a training sample from the minority group $s_{min}$ is made equal to that from the majority group $s_{maj}$: such that $P(\mathbf{x}_i | s_i = s_{min}) = P(\mathbf{x}_{i} | s_i = s_{maj})$. Oversampling is performed by duplicating minority demographic group samples randomly until each demographic group has equal number of training samples. 


\textbf{Pre-processing: Resampling}. In resampling, the sampling probability of each sample $\mathbf{x}_i$ for training is modified and determined by a weight, $w_{s_i}$, which is inversely proportional to the number of samples in $s_i$ \cite{reweight}:
%
%
\begin{equation}\label{eq:weight}
w_{s_i} = \frac{N}{N_{s_i}},
\end{equation}
where $N_{s_i}$ is the number of samples with sensitive attribute $s_i$.



\textbf{In-processing: Loss Reweighting}. 
Loss reweighting is a common method wherein we increase the loss value for a minority group with respect to the inverse of the sample size, which leads to individual minority samples contributing more to training \cite{fairness_class_weight}: 
\begin{equation}
    \frac{1}{N}\sum_{i=1}^{N} w_{s_i} \mathcal{L}(\mathbf{x}_i, \mathbf{y}_i,\hat{\mathbf{y}}_i),
\end{equation}
where $w_{s_i}$ is the weight for the demographic group $s_i$ calculated as in Eq. \ref{eq:weight}.

\subsection{Performance and Fairness Measures}

\noindent\textbf{Performance Measure.} To quantify the performances of the models, we take the average of the angular errors (defined in Eq. \ref{eq:error}) for different demographic groups: $\tilde{\ell}_s = \textrm{avg}(\{\ell_i | {s_i=s}\})$ and $\tilde{\ell}_{s'}=\textrm{avg}(\{\ell_i | {s_i=s'}\})$. 

\noindent\textbf{Fairness Measures.} To measure fairness, angular error distributions $\{\ell_i | {s_i=s}\}$ and $\{\ell_i | {s_i=s'}\}$ are compared via the following measures commonly used in analyzing fairness in regression problems \cite{fair_regression_with_wass_kolmogorov_metric,optimal_transport_wasserstein_metric}:



\noindent\textbf{Fairness Measure: Wasserstein Distance.} Similar to Silvia et al. \cite{fair_regression_with_wass_kolmogorov_metric,optimal_transport_wasserstein_metric}, we calculate the Wasserstein distance between demographic group errors. The intuition is to calculate the total transportation cost to make the error distributions same and use this cost as the amount of dissimilarity:
\begin{equation}
W(s,s')
=
\int_{-\infty}^{+\infty}
\left|
E_{s}(\tau)-E_{s'}(\tau)
\right|
\,\mathrm{d}\tau,
\label{eq:wasserstein_1d_cdf}
\end{equation}
and the cumulative distribution function is calculated as follows:
\begin{equation}\label{eq:error_dist}
E_{s}(\tau)
=
\frac{1}{N_s}\sum_{i=1}^{N_s}\mathbf{1}\!\left(\ell_i\le \tau \;\land\; s_i=s\right).
\end{equation}
The model is deemed perfectly fair when $W(s, s')=0$.

\noindent\textbf{Fairness Measure: Kolmogorov–Smirnov Distance.}
Similar to Chzhen et al. \cite{fair_regression_with_wass_kolmogorov_metric}, we  calculate the Kolmogorov-Smirnov (KS) distance between demographic group errors, which is based on the maximum difference between the error distributions:
\begin{equation}
KS(s,s')
=
\sup_{\tau \in \mathbb{R}}
\left|
E_{s}(\tau)
-
E_{s'}(\tau)
\right|,
\end{equation}
where the same cumulative distribution function $E_{s}(\tau)$ is as defined in Eq. \ref{eq:error_dist}. The model is deemed perfectly fair when $KS(s, s')=0.$

\noindent\textbf{Fairness Measure: $t$- and $p$-values.} Following prior work \cite{novel_approach_t_test_fairness_metric}, we use $t$ and $p$ values between the error distributions to quantify fairness. The $t$ and $p$ values are calculated by using the two sample $t$-test \cite{t_test_two_sample} to check whether the difference between errors of two groups are statistically significant, i.e. if they follow a similar distribution. The two-sample $t$-test is calculated with:
\begin{equation}
t
=
\frac{\tilde \ell_{s} - \tilde \ell_{s'}}
{\sqrt{
\left(\frac{1}{N_s}+\frac{1}{N_{s'}}\right)
\frac{(N_s-1)\sigma_{s}^{2} + (N_{s'}-1)\sigma_{s'}^{2}}
{N_s+N_{s'}-2}
}} .
\end{equation}
where $\tilde \ell_{s}$ and $\sigma_{s}$ are the error mean and variances for the sensitive attribute $s$, respectively. The $p$ value is calculated as:
\begin{equation}
p = P\left(T_{df} \ge t\right),
\end{equation}
with $T_{df}$ denoting a random value from t-distribution with $df=N_s+N_{s'}+1$ degrees of freedom.




\section{Experiments and Results}

In the experiments, we seek to answer the following research questions: 

\begin{itemize}
    \item RQ 1: Are existing gaze datasets biased across demographic groups?
    \item RQ 2: Are appearance-based gaze estimation models biased in terms of their gaze predictions across different demographic groups? 
    \item RQ 3: How effective are bias mitigation methods commonly used in regression problems effective against the biases discovered in RQ 1?
\end{itemize}

Experiments 1, 2 and 3 respectively focus on answering RQs 1, 2 and 3.

\subsection{Implementation and Training Details}

\noindent\textbf{Evaluated Models.} For evaluation, we selected the state-of-the-art gaze estimation models with publicly available implementations. These include CrossGaze \cite{crossgaze}, MCGaze \cite{mcgaze}, L2CS-Net \cite{l2csnet}, PureGaze \cite{puregaze}, and GazeTR \cite{gazetr}. 

\noindent\textbf{Datasets.} We use the commonly-used gaze estimation datasets Gaze360 \cite{gaze360} and GazeCapture \cite{gazecapture}, which we annotated with sensitive attributes (see Section \ref{sect:data_annotation}).

\noindent\textbf{Training Details.} The selected models are trained on Gaze360 dataset for all experiments. The same training hyperparameters are adopted as implemented by the original authors and we ensured that we obtained comparable results with the original papers. In oversampling experiments, we adjust the learning rate ($\eta$) proportionally between the ratio of original ($\mathcal{D}$) and oversampled dataset ($\mathcal{D}^+$) sizes as follows (following \cite{square_root_scaling_rule}):
\begin{equation}
\eta_{\mathcal{+}}
=
\eta
\cdot
\sqrt{
\frac
{|\mathcal{D}|}
{|\mathcal{D^{+}}|}.
}
\end{equation}

\noindent\textbf{Implementation Details.} Frame-based angular errors are calculated using Eq. \ref{eq:error}. The reported fairness metrics are calculated on the frame-based angular errors. Experiments are conducted on Gaze360 and GazeCapture datasets with and without applying the bias mitigation methods described in Section \ref{sect:bias_mitigation}. We exclude the ``Unsure" ethnicity-labeled samples from analysis due to their label ambiguity.




\begin{table}[hbt!]
\setlength{\tabcolsep}{1.8pt}
\renewcommand{\arraystretch}{1.2}
\caption{Exp 1: The distribution of samples across demographic groups for the two datasets. Cau: Caucasian, Afro-A: Afro-American.
    \label{tab:dataset_distribution}}
    \centering\scriptsize
    \begin{tabular}{|c|cccc|cc|c|} \hline 
                & \multicolumn{4}{c|}{Ethnicity} &  \multicolumn{2}{c|}{Gender} & \\
      Dataset   &  Cau. & Afro-A. & Asian & Unsure & Male  & Female & Total \\ \hline \hline 
      Gaze360   & 118,650 & 12,950 & 64,041 & 1,947 & 103,805 & 93,783 & 197,588 \\   \hline
      GazeCapture   & 1,827,875 & 178,250 & 241,127 & 198,252 & 1,244,510 & 1,200,994 & 2,445,504 \\   \hline
    \end{tabular}
    
\end{table}

\subsection{Exp 1: Analysis of Dataset Bias in Gaze Datasets}

In this experiment, we analyze RQ 1 by inspecting the distribution of sensitive attributes for the two datasets. Looking at the distribution of samples in Table \ref{tab:dataset_distribution}, we observe that the cardinalities of Afro-American and Asian samples are substantially lower than that of Caucasian for both datasets. As for the gender sensitive attribute, we note that male and female samples are comparable in size for both datasets. 

\setlength\FrameSep{\fboxsep}
\begin{framed}
\noindent\textbf{Key Takeaway:} Experiment 1 answers RQ 1 by showing that gaze estimation datasets include severe bias across ethnicity groups. However, we do not observe bias across gender in the datasets.
\end{framed}

\subsection{Exp 2: Analysis of Bias in Gaze Estimation}

With this experiment, we analyze RQ 2 separately for the two dataset.


\textbf{Gaze360}:
With reference to Table \ref{tab:reproduce_gaze360_ethnicity}, the analysis results suggest that the models exhibit biased outcomes. The strongest bias appears to be for the Caucasian -- Afro-American ethnicity pair, as indicated by the largest WS and KS values as well as the largest $|t|$ and lowest $p$ values across all models. The comparison between the predictions on Asian -- Caucasian samples also reveals strong bias, with the exception of PureGaze being fair among the models. The models appear fairer across Afro-American and Asian predictions overall. 
Across the gender attribute, we observe fairer predictions for L2CS-Net and PureGaze models. Moreover, the fairness measures overall indicate less disparity compared to the ethnicity groups. 


\textbf{GazeCapture}: The results in Table \ref{tab:reproduce_gazecapture_ethnicity} indicate biased predictions for the GazeCapture dataset too. Similar to the Gaze360 dataset, the comparison between samples of different ethnicities reveal strong biases especially towards the Afro-American samples (as indicated by largest $|t|$ values). With the exception of PureGaze, which produces fair predictions between Asian and Caucasian samples, all models exhibit different levels of biases across ethnicity labels. As for the gender attribute, we notice CrossGaze and L2CS-Net in their predictions. However, the other models suffer from biased outcomes in general. 

We note that the models are trained on Gaze360 but tested on GazeCapture. This deliberate evaluation targets analyzing the generalization performances of the methods across different demographic groups. The results indicate that some models (such as PureGaze and GazeTR) are relatively robust against dataset change whereas some (e.g., MCGaze) are severely affected by the change. 


\input{Tables/exp2_analysis_baseline_results}

\setlength\FrameSep{\fboxsep}
\begin{framed}
\noindent\textbf{Key Takeaway:} Experiment 2 answers RQ 2 by showing that appearance-based gaze estimation models produce biased outcomes across both ethnicity and gender. 
\end{framed}

\subsection{Exp 3: Mitigating Bias in Gaze Estimation}

In this experiment, we evaluate the effectiveness of bias mitigation strategies on gaze estimation. In the following, we report the results across tables to facilitate scrutiny. For a visualization of the numerical values, see the supplementary material.

\noindent\textbf{Pre-processing: Oversampling Results.} The results in Table \ref{tab:Oversample_table_result} indicate that oversampling leads to consistent improvement for PureGaze across all measures and all ethnicity and gender comparisons on Gaze360. However, this is not consistently observed for the the other models across ethnicity on both datasets. In fact, for CrossGaze across Afro-American-Asian and Caucasian -- Afro-American comparisons, oversampling worsens the gaze estimation errors as well as the fairness measures on Gaze360. We observe similar inconsistent behaviors for the gender attribute: Oversampling does not consistently help with fairness.  

\noindent\textbf{Pre-processing: Resampling Results.} Table \ref{tab:Resampling_table_result} lists the performances of the methods when resampling is applied. The results highlight that PureGaze consistently benefits from resampling across ethnicity as well as gender labels on Gaze360. On the GazeCapture dataset, though, no model exhibits consistent improvement across demographic groups. 


\noindent\textbf{In-processing: Loss Reweighting.} The results in Table \ref{tab:loss_reweighing_table_result} point to a similar pattern as the other bias mitigation methods: We do not observe consistent improvements in fairness across different demographic attributes and models. However, $|t|$ values indicate that PureGaze provides fairer predictions across ethnicity and gender attributes for the Gaze360 dataset. 

\noindent\textbf{Which method is the most effective?} We compare the effectiveness of different bias mitigation strategies by counting the number of times they lead to an improvement in a fairness measure on a dataset. The results in Table \ref{tab:which_is_best} suggest that resampling and loss reweighting methods provide the the most number of improvements across models.

\input{Tables/exp3_bias_mitigation_oversample_results}

\input{Tables/exp3_bias_mitigation_resampling_results}

\input{Tables/exp3_bias_mitigation_lossReweighting_results}

\begin{table}[]\renewcommand{\arraystretch}{1.2}
    \centering
    \caption{Exp 3: The number of times each method improves fairness across any of the measures. Each improvement on a dataset with respect to a measure is counted independently as one (1).  
    \label{tab:which_is_best}}
    \begin{tabular}{lcccccc} \hline
       Model & Oversampling & Resampling & Loss Reweighting \\ \hline
       CrossGaze & 12 & 14 & 19 \\
       MCGaze & 9 & 11 & 14 \\
       L2CS-Net & 16 & 16 &  18\\
       PureGaze & 18 & 23 &  16\\
       GazeTR & 13 & 13 &  10\\ \hline
       Total & 68 & \textbf{77} & \textbf{77} \\ \hline        
    \end{tabular}
\end{table}

\setlength\FrameSep{\fboxsep}
\begin{framed}
\noindent\textbf{Key Takeaway:} Experiment 3 answers RQ 3 by demonstrating the effectiveness of different methods in mitigating biases of appearance-based gaze estimation models. We especially observe resampling and loss reweighting are the most effective methods in mitigating biases. However, we notice that methods do not consistently improve fairness across different sensitive attributes or datasets. We discuss possible reasons in Discussion.
\end{framed}

\section{Conclusion and Discussion}

\subsection{Summary} 

In this paper, for the first time, we investigated bias in gaze estimation datasets and fairness of the state-of-the-art gaze estimation models across different demographic groups To be able to do so, we annotated sensitive labels (ethnicity and gender) for the commonly used gaze estimation datasets. Furthermore, we applied commonly-used bias mitigation methods with our annotations and evaluated their effectiveness. 

Overall, our paper highlights bias and fairness as an important concern for appearance-based gaze estimation models, which has been neglected in the gaze estimation literature so far. Moreover, it provides baseline performances of common bias mitigation strategies, which suggest that there is large space for improvement. Our work calls for for further research into the design of fairer gaze estimation models and more effective bias mitigation methods.


\subsection{Discussion}

\noindent\textbf{Key take-aways.} We highlight the following as the key messages of our results:

\begin{itemize}
    \item Appearance-based gaze estimation methods suffer from significant fairness issues across different demographic groups. For example, Caucasian -- Afro-American ethnicity pair demonstrates bias for all experiments.  
    
    \item Existing bias mitigation methods help with fairness to a certain degree. We do not observe consistent fairness improvements by any method across different models. 

    \item Some bias mitigation methods methods even increased bias and removed fair cases. Furthermore, some methods achieved fairness while decreasing gaze estimation accuracy.
    
    \item We observe that resampling and loss reweighting  are the most effective in improving fairness of existing models.
\end{itemize}

\noindent\textbf{On the Limited Effectiveness of Bias Mitigation Methods.} Our results suggest that the considered bias mitigation methods were limited in their effectiveness in improving fairness. Possible reasons for these results include: 

\begin{itemize} 

\item Methods like oversampling address imbalance (quantity) but do not address the inherent `observability' differences (quality) in the data. If gaze features are less distinct in certain demographic groups due to lower contrast or lighting conditions, simply increasing the weight of these samples forces the model to overfit to noise rather than learning generalizable gaze features.

\item Gaze signals are highly entangled with appearance attributes. Simple reweighting strategies operate on the loss function but may fail to disentangle the gaze signal from ethnicity-specific appearance features deep within the network backbone. The model may effectively be learning to predict 'ethnicity' as a proxy for gaze corrections rather than learning true gaze geometry.

\item Bias is not solely demographic but also contextual. If specific demographic groups in the training data are correlated with specific lighting conditions or recording environments (confounding variables), rebalancing the demographics does not remove the environmental bias the model relies on.

\end{itemize}

\noindent\textbf{Open Issues and Call for Action.} The current gaze estimation datasets do not represent different ethnicities and genders equally. To achieve fairer models, ethnicity and gender balanced datasets can be collected, annotated, and used. However, note that balanced number of samples across demographic groups may not guarantee fairness \cite{cheong2023towards,wang2019balanced}. 

Additionally, more advanced bias mitigation methods can be applied to gaze estimation models. In the literature, there is limited work considering in-processing and postprocessing bias mitigation methods in regression setting for facial signal processing domain, and to our knowledge, no comprehensive work is applied on gaze estimation task.

Furthermore, existing post-processing methods for regression problems consider model outputs as well as sensitive attributes to calibrate model outputs to reduce bias. However, in the case of estimated gaze directions, such an approach is not promising since the input facial image is not considered. To the best of our knowledge, such a method has not been proposed in the literature.

Finally, fairness is generally studied for classification tasks, and subsequently, there is very little work on regression-based fairness metrics in the literature. Fairness metrics in regression and especially, for gaze estimation, should be studied more extensively.

\noindent\textbf{Limitations and Future Work.} Our work forms a baseline for bias and fairness in appearance-based gaze estimation. Although we considered commonly used models and bias mitigation methods, our sensitive attribute annotations and comparisons opens the way for further research using more estimation models and bias mitigation methods. Furthermore, it is promising to explore bias and fairness across age, which can provide complementary insights to our work.





\section*{Acknowledgment}

This work was supported by the Council of Higher Education Research Universities Support program through METU Scientific Research Projects  (Project No. ADEP-312-2024-11469) and TÜBİTAK (Project no: 120E269). 

{\small
\bibliographystyle{ieee}
\bibliography{egbib}
}

\end{document}

%% file: Tables/exp2_analysis_baseline_results.tex
\begin{table}[h!]
\caption{Exp 2: Analysis of bias in existing gaze estimation models. WS and KS do not have accepted thresholds to identify fairness whereas $1-p \leq 0.95$ indicates fairness. \textcolor{darkred}{Red} and \textcolor{darkgreen}{green} numbers indicate bias and fairness, respectively. $^*$ denotes models that are trained on Gaze360 dataset and tested on GazeCapture dataset. To make it easier to follow the tables, we make all fairness metrics lower-better by converting the $p$-value metric and its threshold ($p>0.05$) into $1-p \leq 0.95$.}
    \centering
    \label{tab:reproduce_table_result}
    
\renewcommand{\arraystretch}{1.1}
\begin{subtable}{0.49\textwidth}\centering
\caption{Gaze360 dataset results.}
\label{tab:reproduce_gaze360_ethnicity}
\setlength{\tabcolsep}{5pt}
\begin{tabular}{lcccccc}
\hline \\[-7pt] 
Model & $\tilde{\ell}_s$ $\downarrow$ & $\tilde{\ell}_{s'}$ $\downarrow$ & WS $\downarrow$ & KS $\downarrow$ & $|t|$ $\downarrow$ & $1-p$ $\downarrow$ \\ 
\hline
\rowcolor{gray!10}\multicolumn{7}{c}{\textit{Asian -- Caucasian ethnicity pair results.}}\\ \hline

CrossGaze & 10.213 & 9.840 & 0.601 & 0.026 & 2.683 & \textcolor{darkred}{0.993} \\
MCGaze & 11.259 & 10.253 & 1.016 & 0.056 & 6.919 & \textcolor{darkred}{1.000} \\
L2CS-Net & 10.962 & 10.387 & 0.593 & 0.024 & 4.050 & \textcolor{darkred}{1.000} \\
PureGaze & 11.056 & 10.714 & 0.414 & 0.029 & 1.724 & \textcolor{darkgreen}{0.915} \\
GazeTR & 11.362 & 10.178 & 1.185 & 0.047 & 6.945 & \textcolor{darkred}{1.000} \\
\hline
\rowcolor{gray!10}\multicolumn{7}{c}{\textit{Afro-American -- Asian ethnicity pair results.}}\\ \hline

CrossGaze & 10.686 & 10.213 & 0.669 & 0.067 & 1.991 & \textcolor{darkred}{0.954} \\
MCGaze & 11.725 & 11.259 & 0.948 & 0.110 & 1.785 & \textcolor{darkgreen}{0.926} \\
L2CS-Net & 11.759 & 10.962 & 0.834 & 0.097 & 3.217 & \textcolor{darkred}{0.999} \\
PureGaze & 12.253 & 11.056 & 1.248 & 0.094 & 3.629 & \textcolor{darkred}{1.000} \\
GazeTR & 11.720 & 11.362 & 0.745 & 0.058 & 1.149 & \textcolor{darkgreen}{0.750} \\
\hline
\rowcolor{gray!10}\multicolumn{7}{c}{\textit{Caucasian -- Afro-American ethnicity pair results.}}\\ \hline

CrossGaze & 9.840 & 10.686 & 0.847 & 0.056 & 4.522 & \textcolor{darkred}{1.000} \\
MCGaze & 10.253 & 11.725 & 1.504 & 0.119 & 7.868 & \textcolor{darkred}{1.000} \\
L2CS-Net & 10.387 & 11.759 & 1.376 & 0.095 & 7.053 & \textcolor{darkred}{1.000} \\
PureGaze & 10.714 & 12.253 & 1.550 & 0.116 & 5.614 & \textcolor{darkred}{1.000} \\
GazeTR & 10.178 & 11.720 & 1.544 & 0.088 & 6.895 & \textcolor{darkred}{1.000} \\
\hline
\rowcolor{gray!10}\multicolumn{7}{c}{\textit{Male -- Female gender pair results.}}\\ \hline

CrossGaze & 10.364 & 9.915 & 0.495 & 0.039 & 3.324 & \textcolor{darkred}{0.999} \\
MCGaze & 11.254 & 10.494 & 0.846 & 0.088 & 5.374 & \textcolor{darkred}{1.000} \\
L2CS-Net & 10.826 & 10.609 & 0.430 & 0.024 & 1.553 & \textcolor{darkgreen}{0.879} \\
PureGaze & 11.262 & 10.881 & 0.643 & 0.041 & 1.962 & \textcolor{darkgreen}{0.950} \\
GazeTR & 11.121 & 10.491 & 0.662 & 0.036 & 3.778 & \textcolor{darkred}{1.000} \\
\hline
\end{tabular}
\end{subtable}

\renewcommand{\arraystretch}{1.1}
\begin{subtable}{0.49\textwidth} \centering
\vspace*{0.2cm}\caption{GazeCapture dataset results.}
\label{tab:reproduce_gazecapture_ethnicity}
\setlength{\tabcolsep}{5pt}
\begin{tabular}{lcccccc}
\hline \\[-7pt] 
Model & $\tilde{\ell}_s$ $\downarrow$ & $\tilde{\ell}_{s'}$ $\downarrow$ & WS $\downarrow$ & KS $\downarrow$ & $|t|$ $\downarrow$ & $1-p$ $\downarrow$ \\
\hline
\rowcolor{gray!10}\multicolumn{7}{c}{\textit{Asian -- Caucasian ethnicity pair results.}}\\ \hline

CrossGaze$^*$ & 17.598 & 16.790 & 1.172 & 0.047 & 10.757 & \textcolor{darkred}{1.000} \\
MCGaze$^*$ & 41.751 & 40.816 & 2.757 & 0.034 & 3.968 & \textcolor{darkred}{1.000} \\
L2CS-Net$^*$ & 17.516 & 18.102 & 1.114 & 0.055 & 7.329 & \textcolor{darkred}{1.000} \\
PureGaze$^*$ & 16.442 & 16.372 & 0.768 & 0.028 & 0.955 & \textcolor{darkgreen}{0.661} \\
GazeTR$^*$ & 16.560 & 19.509 & 2.949 & 0.061 & 24.652 & \textcolor{darkred}{1.000} \\
\hline
\rowcolor{gray!10}\multicolumn{7}{c}{\textit{Afro-American -- Asian ethnicity pair results.}}\\ \hline

CrossGaze$^*$ & 21.157 & 17.598 & 3.560 & 0.161 & 28.246 & \textcolor{darkred}{1.000} \\
MCGaze$^*$ & 37.703 & 41.751 & 4.253 & 0.059 & 11.303 & \textcolor{darkred}{1.000} \\
L2CS-Net$^*$ & 19.719 & 17.516 & 2.204 & 0.107 & 17.244 & \textcolor{darkred}{1.000} \\
PureGaze$^*$ & 18.575 & 16.442 & 2.133 & 0.107 & 17.704 & \textcolor{darkred}{1.000} \\
GazeTR$^*$ & 19.953 & 16.560 & 3.397 & 0.151 & 27.256 & \textcolor{darkred}{1.000} \\
\hline
\rowcolor{gray!10}\multicolumn{7}{c}{\textit{Caucasian -- Afro-American ethnicity pair results.}}\\ \hline

CrossGaze$^*$ & 16.790 & 21.157 & 4.368 & 0.170 & 49.953 & \textcolor{darkred}{1.000} \\
MCGaze$^*$ & 40.816 & 37.703 & 3.145 & 0.066 & 11.505 & \textcolor{darkred}{1.000} \\
L2CS-Net$^*$ & 18.102 & 19.719 & 1.617 & 0.055 & 17.347 & \textcolor{darkred}{1.000} \\
PureGaze$^*$ & 16.372 & 18.575 & 2.203 & 0.083 & 25.830 & \textcolor{darkred}{1.000} \\
GazeTR$^*$ & 19.509 & 19.953 & 2.581 & 0.094 & 3.154 & \textcolor{darkred}{0.998} \\
\hline
\rowcolor{gray!10}\multicolumn{7}{c}{\textit{Male -- Female gender pair results.}}\\ \hline

CrossGaze$^*$ & 17.328 & 17.241 & 0.203 & 0.009 & 1.826 & \textcolor{darkgreen}{0.932} \\
MCGaze$^*$ & 39.549 & 42.098 & 2.988 & 0.066 & 17.253 & \textcolor{darkred}{1.000} \\
L2CS-Net$^*$ & 18.115 & 18.121 & 0.175 & 0.010 & 0.130 & \textcolor{darkgreen}{0.104} \\
PureGaze$^*$ & 16.692 & 16.351 & 0.345 & 0.021 & 7.418 & \textcolor{darkred}{1.000} \\
GazeTR$^*$ & 18.793 & 19.366 & 0.976 & 0.031 & 7.956 & \textcolor{darkred}{1.000} \\
\hline
\end{tabular}
\end{subtable}
\end{table}

%% file: Tables/exp3_bias_mitigation_oversample_results.tex
\begin{table}[h!]
\caption{Exp 3: Bias mitigation results for \textbf{oversampling} experiment. WS and KS do not have accepted thresholds to identify fairness whereas $1-p \leq 0.95$ indicates fairness. \textcolor{darkred}{Red} and \textcolor{darkgreen}{green} numbers indicate bias and fairness, respectively. $\updegrade$, $\downimprove$, $\nochange$ denote respectively improvement,  reduction or no-change in fairness compared to Table \ref{tab:reproduce_table_result}. $^*$ denotes models that are trained on Gaze360 dataset and tested on GazeCapture dataset. To make it easier to follow the tables, we make all fairness metrics lower-better by converting the $p$-value metric and its threshold ($p>0.05$) into $1-p \leq 0.95$. \label{tab:Oversample_table_result}}
\renewcommand{\arraystretch}{1.05}
\begin{subtable}{0.45\textwidth}
\caption{Gaze360 dataset results. \label{tab:oversample_gaze360}}
\setlength{\tabcolsep}{3.5pt}
\begin{tabular}{lcccccc}
\hline \\[-7pt] 
Model & $\tilde{\ell}_s$ $\downarrow$ & $\tilde{\ell}_{s'}$ $\downarrow$ & WS $\downarrow$ & KS $\downarrow$ & $|t|$ $\downarrow$ & $1-p$ $\downarrow$ \\
\hline
\rowcolor{gray!10}\multicolumn{7}{c}{\textit{Asian -- Caucasian ethnicity pair results.}}\\ 

CrossGaze & 10.305$\updegrade$ & 9.960$\updegrade$ & 0.620$\updegrade$ & 0.026$\nochange$ & 2.466$\downimprove$ & \textcolor{darkred}{0.986$\downimprove$} \\
MCGaze & 11.729$\updegrade$ & 10.137$\downimprove$ & 1.601$\updegrade$ & 0.095$\updegrade$ & 11.039$\updegrade$ & \textcolor{darkred}{1.000}$\nochange$ \\
L2CS-Net & 10.730$\downimprove$ & 10.407$\updegrade$ & 0.447$\downimprove$ & 0.020$\downimprove$ & 2.234$\downimprove$ & \textcolor{darkred}{0.974$\downimprove$} \\
PureGaze & 10.912$\downimprove$ & 10.654$\downimprove$ & 0.390$\downimprove$ & 0.022$\downimprove$ & 1.275$\downimprove$ & \textcolor{darkgreen}{0.798$\downimprove$} \\
GazeTR & 11.443$\updegrade$ & 10.261$\updegrade$ & 1.186$\updegrade$ & 0.050$\updegrade$ & 6.999$\updegrade$ & \textcolor{darkred}{1.000}$\nochange$ \\
\hline
\rowcolor{gray!10}\multicolumn{7}{c}{\textit{Afro-American -- Asian ethnicity pair results.}}\\ 

CrossGaze & 11.456$\updegrade$ & 10.305$\updegrade$ & 1.151$\updegrade$ & 0.094$\updegrade$ & 4.730$\updegrade$ & \textcolor{darkred}{1.000$\updegrade$} \\
MCGaze & 12.258$\updegrade$ & 11.729$\updegrade$ & 0.652$\downimprove$ & 0.075$\downimprove$ & 1.962$\updegrade$ & \textcolor{darkred}{0.950$\updegrade$} \\
L2CS-Net & 11.620$\downimprove$ & 10.730$\downimprove$ & 0.902$\updegrade$ & 0.058$\downimprove$ & 3.544$\updegrade$ & \textcolor{darkred}{1.000$\updegrade$} \\
PureGaze & 11.796$\downimprove$ & 10.912$\downimprove$ & 0.902$\downimprove$ & 0.063$\downimprove$ & 2.619$\downimprove$ & \textcolor{darkred}{0.991$\downimprove$} \\
GazeTR & 11.679$\downimprove$ & 11.443$\updegrade$ & 0.696$\downimprove$ & 0.048$\downimprove$ & 0.773$\downimprove$ & \textcolor{darkgreen}{0.561$\downimprove$} \\
\hline
\rowcolor{gray!10}\multicolumn{7}{c}{\textit{Caucasian -- Afro-American ethnicity pair results.}}\\

CrossGaze & 9.960$\updegrade$ & 11.456$\updegrade$ & 1.496$\updegrade$ & 0.074$\updegrade$ & 7.845$\updegrade$ & \textcolor{darkred}{1.000}$\nochange$ \\
MCGaze & 10.137$\downimprove$ & 12.258$\updegrade$ & 2.146$\updegrade$ & 0.121$\updegrade$ & 11.181$\updegrade$ & \textcolor{darkred}{1.000}$\nochange$ \\
L2CS-Net & 10.407$\updegrade$ & 11.620$\downimprove$ & 1.216$\downimprove$ & 0.052$\downimprove$ & 6.030$\downimprove$ & \textcolor{darkred}{1.000}$\nochange$ \\
PureGaze & 10.654$\downimprove$ & 11.796$\downimprove$ & 1.147$\downimprove$ & 0.078$\downimprove$ & 4.072$\downimprove$ & \textcolor{darkred}{1.000}$\nochange$ \\
GazeTR & 10.261$\updegrade$ & 11.679$\downimprove$ & 1.420$\downimprove$ & 0.092$\updegrade$ & 6.327$\downimprove$ & \textcolor{darkred}{1.000}$\nochange$ \\
\hline
\rowcolor{gray!10}\multicolumn{7}{c}{\textit{Male -- Female gender pair results.}}\\ 

CrossGaze & 10.825$\updegrade$ & 10.356$\updegrade$ & 0.534$\updegrade$ & 0.037$\downimprove$ & 3.359$\updegrade$ & \textcolor{darkred}{0.999}$\nochange$ \\
MCGaze & 11.540$\updegrade$ & 10.485$\downimprove$ & 1.145$\updegrade$ & 0.102$\updegrade$ & 7.339$\updegrade$ & \textcolor{darkred}{1.000}$\nochange$ \\
L2CS-Net & 10.766$\downimprove$ & 10.447$\downimprove$ & 0.417$\downimprove$ & 0.021$\downimprove$ & 2.286$\updegrade$ & \textcolor{darkred}{0.978$\updegrade$} \\
PureGaze & 11.035$\downimprove$ & 10.763$\downimprove$ & 0.402$\downimprove$ & 0.030$\downimprove$ & 1.369$\downimprove$ & \textcolor{darkgreen}{0.829$\downimprove$} \\
GazeTR & 11.184$\updegrade$ & 10.529$\updegrade$ & 0.694$\updegrade$ & 0.039$\updegrade$ & 3.977$\updegrade$ & \textcolor{darkred}{1.000}$\nochange$ \\
\hline
\end{tabular}
\end{subtable}
\renewcommand{\arraystretch}{1.05}
\begin{subtable}{0.45\textwidth}
\vspace*{0.2cm} \caption{GazeCapture dataset results.}
\label{tab:oversample_gazecapture}
\setlength{\tabcolsep}{3pt}
\begin{tabular}{lcccccc}
\hline \\[-7pt] 
Model & $\tilde{\ell}_s$ $\downarrow$ & $\tilde{\ell}_{s'}$ $\downarrow$ & WS $\downarrow$ & KS $\downarrow$ & $|t|$ $\downarrow$ & $1-p$ $\downarrow$ \\
\hline
\rowcolor{gray!10}\multicolumn{7}{c}{\textit{Asian -- Caucasian ethnicity pair results.}}\\ 

CrossGaze$^*$ & 17.750$\updegrade$ & 17.555$\updegrade$ & 1.185$\updegrade$ & 0.045$\downimprove$ & 2.539$\downimprove$ & \textcolor{darkred}{0.989$\downimprove$} \\
MCGaze$^*$ & 43.642$\updegrade$ & 43.520$\updegrade$ & 2.260$\downimprove$ & 0.031$\downimprove$ & 0.443$\downimprove$ & \textcolor{darkgreen}{0.342$\downimprove$} \\
L2CS-Net$^*$ & 17.032$\downimprove$ & 17.331$\downimprove$ & 1.003$\downimprove$ & 0.045$\downimprove$ & 3.704$\downimprove$ & \textcolor{darkred}{1.000}$\nochange$ \\
PureGaze$^*$ & 15.932$\downimprove$ & 16.060$\downimprove$ & 0.807$\updegrade$ & 0.034$\updegrade$ & 1.792$\updegrade$ & \textcolor{darkgreen}{0.927$\updegrade$} \\
GazeTR$^*$ & 16.521$\downimprove$ & 20.543$\updegrade$ & 4.022$\updegrade$ & 0.072$\updegrade$ & 29.159$\updegrade$ & \textcolor{darkred}{1.000}$\nochange$ \\
\hline
\rowcolor{gray!10}\multicolumn{7}{c}{\textit{Afro-American -- Asian ethnicity pair results.}}\\ 

CrossGaze$^*$ & 20.759$\downimprove$ & 17.750$\updegrade$ & 3.009$\downimprove$ & 0.136$\downimprove$ & 23.711$\downimprove$ & \textcolor{darkred}{1.000}$\nochange$ \\
MCGaze$^*$ & 36.764$\downimprove$ & 43.642$\updegrade$ & 6.884$\updegrade$ & 0.078$\updegrade$ & 17.669$\updegrade$ & \textcolor{darkred}{1.000} $\nochange$\\
L2CS-Net$^*$ & 19.218$\downimprove$ & 17.032$\downimprove$ & 2.186$\downimprove$ & 0.113$\updegrade$ & 17.206$\downimprove$ & \textcolor{darkred}{1.000}$\nochange$ \\
PureGaze$^*$ & 18.093$\downimprove$ & 15.932$\downimprove$ & 2.161$\updegrade$ & 0.107$\nochange$ & 18.574$\updegrade$ & \textcolor{darkred}{1.000}$\nochange$ \\
GazeTR$^*$ & 19.822$\downimprove$ & 16.521$\downimprove$ & 3.302$\downimprove$ & 0.145$\downimprove$ & 25.642$\downimprove$ & \textcolor{darkred}{1.000}$\nochange$ \\
\hline
\rowcolor{gray!10}\multicolumn{7}{c}{\textit{Caucasian -- Afro-American ethnicity pair results.}}\\ 

CrossGaze$^*$ & 17.555$\updegrade$ & 20.759$\downimprove$ & 3.204$\downimprove$ & 0.117$\downimprove$ & 35.925$\downimprove$ & \textcolor{darkred}{1.000}$\nochange$ \\
MCGaze$^*$ & 43.520$\updegrade$ & 36.764$\downimprove$ & 6.771$\updegrade$ & 0.094$\updegrade$ & 21.475$\updegrade$ & \textcolor{darkred}{1.000}$\nochange$ \\
L2CS-Net$^*$ & 17.331$\downimprove$ & 19.218$\downimprove$ & 1.891$\updegrade$ & 0.070$\updegrade$ & 20.052$\updegrade$ & \textcolor{darkred}{1.000}$\nochange$\\
PureGaze$^*$ & 16.060$\downimprove$ & 18.093$\downimprove$ & 2.033$\downimprove$ & 0.074$\downimprove$ & 24.325$\downimprove$ & \textcolor{darkred}{1.000}$\nochange$ \\
GazeTR$^*$ & 20.543$\updegrade$ & 19.822$\downimprove$ & 2.962$\updegrade$ & 0.080$\downimprove$ & 4.427$\updegrade$ & \textcolor{darkred}{1.000$\updegrade$} \\
\hline
\rowcolor{gray!10}\multicolumn{7}{c}{\textit{Male -- Female gender pair results.}}\\ 

CrossGaze$^*$ & 17.670$\updegrade$ & 18.048$\updegrade$ & 0.474$\updegrade$ & 0.023$\updegrade$ & 7.770$\updegrade$ & \textcolor{darkred}{1.000$\updegrade$} \\
MCGaze$^*$ & 39.693$\updegrade$ & 41.464$\downimprove$ & 2.303$\downimprove$ & 0.055$\downimprove$ & 11.116$\downimprove$ & \textcolor{darkred}{1.000}$\nochange$ \\
L2CS-Net$^*$ & 18.026$\downimprove$ & 17.929$\downimprove$ & 0.201$\updegrade$ & 0.009$\downimprove$ & 1.973$\updegrade$ & \textcolor{darkred}{0.952$\updegrade$} \\
PureGaze$^*$ & 16.229$\downimprove$ & 15.802$\downimprove$ & 0.428$\updegrade$ & 0.028$\updegrade$ & 9.477$\updegrade$ & \textcolor{darkred}{1.000}$\nochange$ \\
GazeTR$^*$ & 18.065$\downimprove$ & 18.304$\downimprove$ & 0.562$\downimprove$ & 0.021$\downimprove$ & 3.842$\downimprove$ & \textcolor{darkred}{1.000}$\nochange$ \\
\hline
\end{tabular}
\end{subtable}
\end{table}

%% file: Tables/exp3_bias_mitigation_resampling_results.tex
\begin{table}[h!]
\caption{Exp 3: Bias mitigation results for \textbf{resampling} experiment. WS and KS do not have accepted thresholds to identify fairness whereas $p \leq 0.95$ indicates fairness.  \textcolor{darkred}{Red} and \textcolor{darkgreen}{green} numbers indicate bias and fairness, respectively. $\updegrade$, $\downimprove$, $\nochange$ denote respectively improvement,  reduction or no-change in fairness compared to Table \ref{tab:reproduce_table_result}. $^*$ denotes models that are trained on Gaze360 dataset and tested on GazeCapture dataset. To make it easier to follow the tables, we make all fairness metrics lower-better by converting the $p$-value metric and its threshold ($p>0.05$) into $1-p \leq 0.95$.}
\label{tab:Resampling_table_result}
\renewcommand{\arraystretch}{1.05}
\begin{subtable}{0.49\textwidth}
\caption{Gaze360 dataset results.}
\label{tab:resample_gaze360_ethnicity}
\setlength{\tabcolsep}{3.9pt}
\begin{tabular}{lcccccc}
\hline \\[-7pt] 
Model &$\tilde{\ell}_s$ $\downarrow$ & $\tilde{\ell}_{s'}$ $\downarrow$ & WS $\downarrow$ & KS $\downarrow$ & $|t|$ $\downarrow$ & $1-p$ $\downarrow$ \\
\hline
\rowcolor{gray!10}\multicolumn{7}{c}{\textit{Asian -- Caucasian ethnicity pair results.}}\\ 

CrossGaze & 10.586$\updegrade$ & 10.062$\updegrade$ & 0.577$\downimprove$ & 0.024$\downimprove$ & 3.666$\updegrade$ & \textcolor{darkred}{1.000$\updegrade$} \\
MCGaze & 11.253$\downimprove$ & 10.136$\downimprove$ & 1.128$\updegrade$ & 0.059$\updegrade$ & 7.634$\updegrade$ & \textcolor{darkred}{1.000}$\nochange$ \\
L2CS-Net & 10.807$\downimprove$ & 10.586$\updegrade$ & 0.412$\downimprove$ & 0.018$\downimprove$ & 1.557$\downimprove$ & \textcolor{darkgreen}{0.881$\downimprove$} \\
PureGaze & 11.199$\updegrade$ & 11.028$\updegrade$ & 0.345$\downimprove$ & 0.020$\downimprove$ & 0.841$\downimprove$ & \textcolor{darkgreen}{0.599$\downimprove$} \\
GazeTR & 11.491$\updegrade$ & 10.242$\updegrade$ & 1.252$\updegrade$ & 0.059$\updegrade$ & 7.449$\updegrade$ & \textcolor{darkred}{1.000}$\nochange$ \\
\hline
\rowcolor{gray!10}\multicolumn{7}{c}{\textit{Afro-American -- Asian ethnicity pair results.}}\\ 

CrossGaze & 11.259$\updegrade$ & 10.586$\updegrade$ & 0.694$\updegrade$ & 0.059$\downimprove$ & 2.701$\updegrade$ & \textcolor{darkred}{0.993$\updegrade$} \\
MCGaze & 11.411$\downimprove$ & 11.253$\downimprove$ & 0.715$\downimprove$ & 0.075$\downimprove$ & 0.608$\downimprove$ & \textcolor{darkgreen}{0.457$\downimprove$} \\
L2CS-Net & 11.646$\downimprove$ & 10.807$\downimprove$ & 0.839$\updegrade$ & 0.059$\downimprove$ & 3.443$\updegrade$ & \textcolor{darkred}{0.999}$\nochange$ \\
PureGaze & 11.687$\downimprove$ & 11.199$\updegrade$ & 0.743$\downimprove$ & 0.063$\downimprove$ & 1.488$\downimprove$ & \textcolor{darkgreen}{0.863$\downimprove$} \\
GazeTR & 11.648$\downimprove$ & 11.491$\updegrade$ & 0.620$\downimprove$ & 0.039$\downimprove$ & 0.522$\downimprove$ & \textcolor{darkgreen}{0.398$\downimprove$} \\
\hline
\rowcolor{gray!10}\multicolumn{7}{c}{\textit{Caucasian -- Afro-American ethnicity pair results.}}\\ 

CrossGaze & 10.062$\updegrade$ & 11.259$\updegrade$ & 1.196$\updegrade$ & 0.057$\updegrade$ & 6.214$\updegrade$ & \textcolor{darkred}{1.000}$\nochange$ \\
MCGaze & 10.136$\downimprove$ & 11.411$\downimprove$ & 1.318$\downimprove$ & 0.105$\downimprove$ & 6.757$\downimprove$ & \textcolor{darkred}{1.000}$\nochange$ \\
L2CS-Net & 10.586$\updegrade$ & 11.646$\downimprove$ & 1.065$\downimprove$ & 0.050$\downimprove$ & 5.418$\downimprove$ & \textcolor{darkred}{1.000}$\nochange$ \\
PureGaze & 11.028$\updegrade$ & 11.687$\downimprove$ & 0.930$\downimprove$ & 0.065$\downimprove$ & 2.340$\downimprove$ & \textcolor{darkred}{0.981$\downimprove$} \\
GazeTR & 10.242$\updegrade$ & 11.648$\downimprove$ & 1.411$\downimprove$ & 0.081$\downimprove$ & 6.291$\downimprove$ & \textcolor{darkred}{1.000}$\nochange$ \\
\hline
\rowcolor{gray!10}\multicolumn{7}{c}{\textit{Male -- female gender pair results.}}\\ 

CrossGaze & 10.570$\updegrade$ & 10.409$\updegrade$ & 0.382$\downimprove$ & 0.024$\downimprove$ & 1.141$\downimprove$ & \textcolor{darkgreen}{0.746$\downimprove$} \\
MCGaze & 11.278$\updegrade$ & 10.477$\downimprove$ & 0.885$\updegrade$ & 0.087$\downimprove$ & 5.589$\updegrade$ & \textcolor{darkred}{1.000}$\nochange$ \\
L2CS-Net & 10.908$\updegrade$ & 10.626$\updegrade$ & 0.340$\downimprove$ & 0.018$\downimprove$ & 2.000$\updegrade$ & \textcolor{darkred}{0.954$\updegrade$} \\
PureGaze & 11.114$\downimprove$ & 10.820$\downimprove$ & 0.487$\downimprove$ & 0.034$\downimprove$ & 1.502$\downimprove$ & \textcolor{darkgreen}{0.867$\downimprove$} \\
GazeTR & 11.212$\updegrade$ & 10.494$\updegrade$ & 0.765$\updegrade$ & 0.039$\updegrade$ & 4.292$\updegrade$ & \textcolor{darkred}{1.000}$\nochange$ \\
\hline
\end{tabular}
\end{subtable}
\renewcommand{\arraystretch}{1.05}
\begin{subtable}{0.45\textwidth}
\vspace*{0.2cm}\caption{GazeCapture dataset results.}
\setlength{\tabcolsep}{3pt}
\begin{tabular}{lcccccc}
\hline \\[-7pt] 
Model & $\tilde{\ell}_s$ $\downarrow$ & $\tilde{\ell}_{s'}$ $\downarrow$ & WS $\downarrow$ & KS $\downarrow$ & $|t|$ $\downarrow$ & $1-p$ $\downarrow$ \\
\hline
\rowcolor{gray!10}\multicolumn{7}{c}{\textit{Asian -- Caucasian ethnicity pair results.}}\\ 

CrossGaze$^*$ & 17.697$\updegrade$ & 17.030$\updegrade$ & 1.180$\updegrade$ & 0.047$\nochange$ & 8.810$\downimprove$ & \textcolor{darkred}{1.000}$\nochange$ \\
MCGaze$^*$ & 39.546$\downimprove$ & 38.372$\downimprove$ & 2.917$\updegrade$ & 0.036$\updegrade$ & 4.980$\updegrade$ & \textcolor{darkred}{1.000}$\nochange$ \\
L2CS-Net$^*$ & 17.605$\updegrade$ & 18.261$\updegrade$ & 1.182$\updegrade$ & 0.061$\updegrade$ & 8.002$\updegrade$ & \textcolor{darkred}{1.000}$\nochange$ \\
PureGaze$^*$ & 16.191$\downimprove$ & 16.272$\downimprove$ & 0.758$\downimprove$ & 0.033$\updegrade$ & 1.121$\updegrade$ & \textcolor{darkgreen}{0.738$\updegrade$} \\
GazeTR$^*$ & 16.715$\updegrade$ & 20.553$\updegrade$ & 3.838$\updegrade$ & 0.071$\updegrade$ & 27.945$\updegrade$ & \textcolor{darkred}{1.000}$\nochange$ \\
\hline
\rowcolor{gray!10}\multicolumn{7}{c}{\textit{Afro-American -- Asian ethnicity pair results.}}\\ 

CrossGaze$^*$ & 20.576$\downimprove$ & 17.697$\updegrade$ & 2.879$\downimprove$ & 0.134$\downimprove$ & 22.730$\downimprove$ & \textcolor{darkred}{1.000}$\nochange$ \\
MCGaze$^*$ & 34.372$\downimprove$ & 39.546$\downimprove$ & 5.208$\updegrade$ & 0.065$\updegrade$ & 14.476$\updegrade$ & \textcolor{darkred}{1.000}$\nochange$ \\
L2CS-Net$^*$ & 19.246$\downimprove$ & 17.605$\updegrade$ & 1.677$\downimprove$ & 0.099$\downimprove$ & 12.790$\downimprove$ & \textcolor{darkred}{1.000}$\nochange$ \\
PureGaze$^*$ & 18.064$\downimprove$ & 16.191$\downimprove$ & 1.873$\downimprove$ & 0.099$\downimprove$ & 16.097$\downimprove$ & \textcolor{darkred}{1.000}$\nochange$ \\
GazeTR$^*$ & 20.576$\updegrade$ & 16.715$\updegrade$ & 3.860$\updegrade$ & 0.164$\updegrade$ & 28.635$\updegrade$ & \textcolor{darkred}{1.000}$\nochange$ \\
\hline
\rowcolor{gray!10}\multicolumn{7}{c}{\textit{Caucasian -- Afro-American ethnicity pair results.}}\\ 

CrossGaze$^*$ & 17.030$\updegrade$ & 20.576$\downimprove$ & 3.546$\downimprove$ & 0.128$\downimprove$ & 40.276$\downimprove$ & \textcolor{darkred}{1.000}$\nochange$ \\
MCGaze$^*$ & 38.372$\downimprove$ & 34.372$\downimprove$ & 4.012$\updegrade$ & 0.068$\updegrade$ & 14.844$\updegrade$ & \textcolor{darkred}{1.000}$\nochange$ \\
L2CS-Net$^*$ & 18.261$\updegrade$ & 19.246$\downimprove$ & 0.986$\downimprove$ & 0.039$\downimprove$ & 10.362$\downimprove$ & \textcolor{darkred}{1.000}$\nochange$ \\
PureGaze$^*$ & 16.272$\downimprove$ & 18.064$\downimprove$ & 1.792$\downimprove$ & 0.071$\downimprove$ & 21.268$\downimprove$ & \textcolor{darkred}{1.000}$\nochange$ \\
GazeTR$^*$ & 20.553$\updegrade$ & 20.576$\updegrade$ & 3.027$\updegrade$ & 0.100$\updegrade$ & 0.138$\downimprove$ & \textcolor{darkgreen}{0.110$\downimprove$} \\
\hline
\rowcolor{gray!10}\multicolumn{7}{c}{\textit{Male -- Female gender pair results.}}\\ 

CrossGaze$^*$ & 17.395$\updegrade$ & 17.542$\updegrade$ & 0.314$\updegrade$ & 0.013$\updegrade$ & 3.052$\updegrade$ & \textcolor{darkred}{0.998$\updegrade$} \\
MCGaze$^*$ & 38.231$\downimprove$ & 40.043$\downimprove$ & 2.219$\downimprove$ & 0.046$\downimprove$ & 12.338$\downimprove$ & \textcolor{darkred}{1.000}$\nochange$ \\
L2CS-Net$^*$ & 18.089$\downimprove$ & 18.082$\downimprove$ & 0.349$\updegrade$ & 0.017$\updegrade$ & 0.133$\updegrade$ & \textcolor{darkgreen}{0.106$\updegrade$} \\
PureGaze$^*$ & 16.411$\downimprove$ & 15.950$\downimprove$ & 0.462$\updegrade$ & 0.025$\updegrade$ & 10.152$\updegrade$ & \textcolor{darkred}{1.000}$\nochange$ \\
GazeTR$^*$ & 18.926$\updegrade$ & 19.154$\downimprove$ & 0.940$\downimprove$ & 0.030$\downimprove$ & 3.473$\downimprove$ & \textcolor{darkred}{0.999$\downimprove$} \\
\hline
\end{tabular}
\end{subtable}
\end{table}

%% file: Tables/exp3_bias_mitigation_lossReweighting_results.tex
\begin{table}[h!]
\caption{Exp 3: Bias mitigation results for \textbf{loss reweighting} experiment. WS and KS do not have accepted thresholds to identify fairness whereas $p \leq 0.95$ indicate fairness. \textcolor{darkred}{Red} and \textcolor{darkgreen}{green} numbers indicate bias and fairness, respectively. $\updegrade$, $\downimprove$, $\nochange$ denote respectively improvement,  reduction or no-change in fairness compared to Table \ref{tab:reproduce_table_result}. $^*$ denotes models that are trained on Gaze360 dataset and tested on GazeCapture dataset. To make it easier to follow the tables, we make all fairness metrics lower-better by converting the $p$-value metric and its threshold ($p>0.05$) into $1-p \leq 0.95$.}
    \centering
    \label{tab:loss_reweighing_table_result}
    
\renewcommand{\arraystretch}{1.05}
\begin{subtable}{0.45\textwidth}\centering
\caption{Gaze360 dataset results.}
\setlength{\tabcolsep}{3.5pt}
\begin{tabular}{lcccccc}
\hline \\[-7pt] 
Model & $\tilde{\ell}_s$ $\downarrow$ & $\tilde{\ell}_{s'}$ $\downarrow$ & WS $\downarrow$ & KS $\downarrow$ & $|t|$ $\downarrow$ & $1-p$ $\downarrow$ \\
\hline 
\rowcolor{gray!10}\multicolumn{7}{c}{\textit{Asian -- Caucasian ethnicity pair results.}}\\ 

CrossGaze & 10.347$\updegrade$ & 9.944$\updegrade$ & 0.652$\updegrade$ & 0.029$\updegrade$ & 2.845$\updegrade$ & \textcolor{darkred}{0.996$\updegrade$} \\
MCGaze & 18.314$\updegrade$ & 18.481$\updegrade$ & 1.503$\updegrade$ & 0.084$\updegrade$ & 0.861$\downimprove$ & \textcolor{darkgreen}{0.611$\downimprove$} \\
L2CS-Net & 10.888$\downimprove$ & 10.605$\updegrade$ & 0.361$\downimprove$ & 0.015$\downimprove$ & 1.956$\downimprove$ & \textcolor{darkgreen}{0.949$\downimprove$} \\
PureGaze & 11.027$\downimprove$ & 10.770$\updegrade$ & 0.404$\downimprove$ & 0.031$\updegrade$ & 1.295$\downimprove$ & \textcolor{darkgreen}{0.805$\downimprove$} \\
GazeTR & 11.406$\updegrade$ & 10.139$\downimprove$ & 1.270$\updegrade$ & 0.055$\updegrade$ & 7.505$\downimprove$ & \textcolor{darkred}{1.000}$\nochange$ \\
\hline
\rowcolor{gray!10}\multicolumn{7}{c}{\textit{Afro-American -- Asian ethnicity pair results.}}\\ 

CrossGaze & 10.693$\updegrade$ & 10.347$\updegrade$ & 0.625$\downimprove$ & 0.064$\downimprove$ & 1.425$\downimprove$ & \textcolor{darkgreen}{0.846$\downimprove$} \\
MCGaze & 16.849$\updegrade$ & 18.314$\updegrade$ & 1.526$\updegrade$ & 0.069$\downimprove$ & 4.530$\updegrade$ & \textcolor{darkred}{1.000$\updegrade$} \\
L2CS-Net & 11.550$\downimprove$ & 10.888$\downimprove$ & 0.665$\downimprove$ & 0.074$\downimprove$ & 2.696$\downimprove$ & \textcolor{darkred}{0.993$\downimprove$} \\
PureGaze & 11.985$\downimprove$ & 11.027$\downimprove$ & 1.008$\downimprove$ & 0.073$\downimprove$ & 2.879$\downimprove$ & \textcolor{darkred}{0.996$\downimprove$} \\
GazeTR & 11.827$\updegrade$ & 11.406$\updegrade$ & 0.796$\updegrade$ & 0.061$\updegrade$ & 1.371$\updegrade$ & \textcolor{darkgreen}{0.830$\updegrade$} \\
\hline
\rowcolor{gray!10}\multicolumn{7}{c}{\textit{Caucasian -- Afro-American ethnicity pair results.}}\\ 

CrossGaze & 9.944$\updegrade$ & 10.693$\updegrade$ & 0.771$\downimprove$ & 0.047$\downimprove$ & 3.941$\downimprove$ & \textcolor{darkred}{1.000}$\nochange$ \\
MCGaze & 18.481$\updegrade$ & 16.849$\updegrade$ & 2.454$\updegrade$ & 0.144$\updegrade$ & 6.317$\downimprove$ & \textcolor{darkred}{1.000}$\nochange$ \\
L2CS-Net & 10.605$\updegrade$ & 11.550$\downimprove$ & 0.947$\downimprove$ & 0.069$\downimprove$ & 4.751$\downimprove$ & \textcolor{darkred}{1.000}$\nochange$ \\
PureGaze & 10.770$\updegrade$ & 11.985$\downimprove$ & 1.249$\downimprove$ & 0.097$\downimprove$ & 4.418$\downimprove$ & \textcolor{darkred}{1.000}$\nochange$ \\
GazeTR & 10.139$\downimprove$ & 11.827$\updegrade$ & 1.689$\updegrade$ & 0.094$\updegrade$ & 7.523$\updegrade$ & \textcolor{darkred}{1.000}$\nochange$ \\
\hline
\rowcolor{gray!10}\multicolumn{7}{c}{\textit{Male -- Female gender pair results.}}\\ 

CrossGaze & 10.683$\updegrade$ & 10.261$\updegrade$ & 0.432$\downimprove$ & 0.026$\downimprove$ & 3.011$\downimprove$ & \textcolor{darkred}{0.997$\downimprove$} \\
MCGaze & 11.203$\downimprove$ & 10.318$\downimprove$ & 0.979$\updegrade$ & 0.094$\updegrade$ & 6.302$\updegrade$ & \textcolor{darkred}{1.000}$\nochange$ \\
L2CS-Net & 10.873$\updegrade$ & 10.517$\downimprove$ & 0.406$\downimprove$ & 0.024$\nochange$ & 2.526$\updegrade$ & \textcolor{darkred}{0.988$\updegrade$} \\
PureGaze & 11.112$\downimprove$ & 10.853$\downimprove$ & 0.422$\downimprove$ & 0.028$\downimprove$ & 1.322$\downimprove$ & \textcolor{darkgreen}{0.814$\downimprove$} \\
GazeTR & 11.067$\downimprove$ & 10.524$\updegrade$ & 0.571$\downimprove$ & 0.030$\downimprove$ & 3.271$\downimprove$ & \textcolor{darkred}{0.999$\downimprove$} \\
\hline
\end{tabular}
\end{subtable}
\renewcommand{\arraystretch}{1.05}
\begin{subtable}{0.45\textwidth}\centering
\vspace*{0.2cm}\caption{GazeCapture dataset results.}
\label{tab:loss_reweighing_gazecapture}
\setlength{\tabcolsep}{3pt}
\begin{tabular}{lcccccc}
\hline \\[-7pt] 
Model & $\tilde{\ell}_s$ $\downarrow$ & $\tilde{\ell}_{s'}$ $\downarrow$ & WS $\downarrow$ & KS $\downarrow$ & $|t|$ $\downarrow$ & $1-p$ $\downarrow$ \\
\hline
\rowcolor{gray!10}\multicolumn{7}{c}{\textit{Asian -- Caucasian ethnicity pair results.}}\\ 

CrossGaze$^*$ & 17.636$\updegrade$ & 16.822$\updegrade$ & 1.197$\updegrade$ & 0.047$\nochange$ & 10.801$\updegrade$ & \textcolor{darkred}{1.000}$\nochange$ \\
MCGaze$^*$ & 27.304$\downimprove$ & 25.704$\downimprove$ & 1.670$\downimprove$ & 0.045$\updegrade$ & 11.999$\updegrade$ & \textcolor{darkred}{1.000}$\nochange$ \\
L2CS-Net$^*$ & 17.817$\updegrade$ & 18.533$\updegrade$ & 1.359$\updegrade$ & 0.067$\updegrade$ & 8.718$\updegrade$ & \textcolor{darkred}{1.000}$\nochange$ \\
PureGaze$^*$ & 16.339$\downimprove$ & 16.129$\downimprove$ & 0.705$\downimprove$ & 0.023$\downimprove$ & 2.920$\updegrade$ & \textcolor{darkred}{0.996$\updegrade$} \\
GazeTR$^*$ & 16.602$\updegrade$ & 18.033$\downimprove$ & 1.431$\downimprove$ & 0.056$\downimprove$ & 15.852$\downimprove$ & \textcolor{darkred}{1.000}$\nochange$ \\
\hline
\rowcolor{gray!10}\multicolumn{7}{c}{\textit{Afro-American -- Asian ethnicity pair results.}}\\ 

CrossGaze$^*$ & 21.144$\downimprove$ & 17.636$\updegrade$ & 3.508$\downimprove$ & 0.155$\downimprove$ & 27.556$\downimprove$ & \textcolor{darkred}{1.000}$\nochange$ \\
MCGaze$^*$ & 25.967$\downimprove$ & 27.304$\downimprove$ & 2.688$\downimprove$ & 0.061$\updegrade$ & 6.582$\downimprove$ & \textcolor{darkred}{1.000}$\nochange$ \\
L2CS-Net$^*$ & 19.360$\downimprove$ & 17.817$\updegrade$ & 1.546$\downimprove$ & 0.085$\downimprove$ & 11.700$\downimprove$ & \textcolor{darkred}{1.000}$\nochange$ \\
PureGaze$^*$ & 18.521$\downimprove$ & 16.339$\downimprove$ & 2.181$\updegrade$ & 0.109$\updegrade$ & 18.277$\updegrade$ & \textcolor{darkred}{1.000}$\nochange$ \\
GazeTR$^*$ & 20.086$\updegrade$ & 16.602$\updegrade$ & 3.484$\updegrade$ & 0.155$\updegrade$ & 28.410$\updegrade$ & \textcolor{darkred}{1.000}$\nochange$ \\
\hline
\rowcolor{gray!10}\multicolumn{7}{c}{\textit{Caucasian -- Afro-American ethnicity pair results.}}\\ 

CrossGaze$^*$ & 16.822$\updegrade$ & 21.144$\downimprove$ & 4.322$\downimprove$ & 0.167$\downimprove$ & 49.261$\downimprove$ & \textcolor{darkred}{1.000}$\nochange$ \\
MCGaze$^*$ & 25.704$\downimprove$ & 25.967$\downimprove$ & 1.314$\downimprove$ & 0.065$\downimprove$ & 1.714$\downimprove$ & \textcolor{darkgreen}{0.913$\downimprove$} \\
L2CS-Net$^*$ & 18.533$\updegrade$ & 19.360$\downimprove$ & 0.849$\downimprove$ & 0.024$\downimprove$ & 8.672$\downimprove$ & \textcolor{darkred}{1.000}$\nochange$ \\
PureGaze$^*$ & 16.129$\downimprove$ & 18.521$\downimprove$ & 2.392$\updegrade$ & 0.090$\updegrade$ & 28.397$\updegrade$ & \textcolor{darkred}{1.000}$\nochange$ \\
GazeTR$^*$ & 18.033$\downimprove$ & 20.086$\updegrade$ & 2.277$\downimprove$ & 0.101$\updegrade$ & 19.371$\updegrade$ & \textcolor{darkred}{1.000$\updegrade$} \\
\hline
\rowcolor{gray!10}\multicolumn{7}{c}{\textit{Male -- Female gender pair results.}}\\ 

CrossGaze$^*$ & 17.349$\updegrade$ & 17.380$\updegrade$ & 0.250$\updegrade$ & 0.009$\nochange$ & 0.643$\downimprove$ & \textcolor{darkgreen}{0.480$\downimprove$} \\
MCGaze$^*$ & 35.198$\downimprove$ & 37.369$\downimprove$ & 2.856$\downimprove$ & 0.057$\downimprove$ & 16.206$\downimprove$ & \textcolor{darkred}{1.000}$\nochange$ \\
L2CS-Net$^*$ & 18.234$\updegrade$ & 18.317$\updegrade$ & 0.192$\updegrade$ & 0.011$\updegrade$ & 1.574$\updegrade$ & \textcolor{darkgreen}{0.885$\updegrade$} \\
PureGaze$^*$ & 16.596$\downimprove$ & 16.219$\downimprove$ & 0.384$\updegrade$ & 0.026$\updegrade$ & 8.187$\updegrade$ & \textcolor{darkred}{1.000}$\nochange$ \\
GazeTR$^*$ & 17.773$\downimprove$ & 18.383$\downimprove$ & 0.680$\downimprove$ & 0.031$\nochange$ & 11.407$\updegrade$ & \textcolor{darkred}{1.000}$\nochange$ \\
\hline
\end{tabular}
\end{subtable}
\end{table}